\documentclass[
]{ceurart}

\usepackage{listings}
\usepackage{tabularx} 

\usepackage{graphicx}
\usepackage{subcaption}

\usepackage{multirow}

\begin{document}

%%
%% Rights management information.
%% CC-BY is default license.
% \copyrightyear{2025}
% \copyrightclause{Copyright for this paper by its authors.
%   Use permitted under Creative Commons License Attribution 4.0
%   International (CC BY 4.0).}

% DO NOT CHANGE THE TEMPLATE BELOW, but only the text
%%
%% This command is for the conference information
% \conference{Late-breaking work, Demos and Doctoral Consortium, colocated with the 3rd World Conference on eXplainable Artificial
% Intelligence: July 09–11, 2025, Istanbul, Turkey}

%%
%% The "title" command
\title{EvalxNLP: A Framework for Benchmarking Post-Hoc Explainability Methods on NLP Models}

% \tnotemark[1]

%%
%% The "author" command and its associated commands are used to define
%% the authors and their affiliations.
\author[]{Mahdi Dhaini}[%
% orcid=0000-0002-7831-3141,
email=mahdi.dhaini@tum.de,
% url=https://yamadharma.github.io/,
]
\cormark[1]
% \fnmark[1]
\address[]{Technical University of Munich, School of Computation, Information and Technology, Department of Computer Science, Boltzmannstr. 3, Garching, 85748, Germany}
% \address[1]{Technical University of Munich, School of Computation, Information and Technology, Department of Computer Science, Arcisstraße 21, Munich, 80333, Germany}

% \address[2]{Joint Institute for Nuclear Research,
%   6 Joliot-Curie, Dubna, Moscow region, 141980, Russian Federation}

\author[]{Kafaite Zahra Hussain}[%
% orcid=0009-0002-1358-2487,
email=ge95nut@mytum.de,
% url=https://kmitd.github.io/ilaria/,
]
% \fnmark[1]
% \address[3]{Vrije Universiteit Amsterdam, De Boelelaan 1105, 1081 HV Amsterdam, The Netherlands}

\author[]{Efstratios Zaradoukas}[%
% orcid=0009-0004-5762-3986,
email=efstratios.zaradoukas@tum.de,
% url=https://www.gov.sot.tum.de/rds/team/stratis-zaradoukas/,
]
% \fnmark[1]
% \address[4]{University of Skövde, Högskolevägen 1, 541 28 Skövde, Sweden}

\author[]{Gjergji Kasneci}[%
% orcid=0000-0002-3123-7268,
email=gjergji.kasneci@tum.de,
% url=http://conceptbase.sourceforge.net/mjf/,
]

%% Footnotes
\cortext[1]{Corresponding author.}
% \fntext[1]{These authors contributed equally.}

%%
%% The abstract is a short summary of the work to be presented in the
%% article.
\begin{abstract}
As Natural Language Processing (NLP) models continue to evolve and become integral to high-stakes applications, ensuring their interpretability remains a critical challenge. Given the growing variety of explainability methods and diverse stakeholder requirements, frameworks that help stakeholders select appropriate explanations tailored to their specific use cases are increasingly important.
To address this need, we introduce \textit{EvalxNLP}, a Python framework for benchmarking state-of-the-art feature attribution methods for transformer-based NLP models. \textit{EvalxNLP} integrates eight widely recognized explainability techniques from the Explainable AI (XAI) literature, enabling users to generate and evaluate explanations based on key properties such as faithfulness, plausibility, and complexity. Our framework also provides interactive, LLM-based textual explanations, facilitating user understanding of the generated explanations and evaluation outcomes. Human evaluation results indicate high user satisfaction with \textit{EvalxNLP}, suggesting it is a promising framework for benchmarking explanation methods across diverse user groups. By offering a user-friendly and extensible platform, \textit{EvalxNLP} aims at democratizing explainability tools and supporting the systematic comparison and advancement of XAI techniques in NLP.
\end{abstract}

%%
%% Keywords. The author(s) should pick words that accurately describe
%% the work being presented. Separate the keywords with commas.

%%
%% This command processes the author and affiliation and title
%% information and builds the first part of the formatted document.
\maketitle

\section{Introduction}
Although significant progress has been made in the field of NLP, transformer-based models often operate as 
\textit{black boxes}, making it challenging to interpret their decision-making processes. In high-stakes domains like medical diagnosis and financial decision-making, transparency is essential for trust and accountability. Despite the rapid development of explainability methods, there remains a lack of standardized evaluation frameworks, particularly for NLP, where text data is inherently unstructured and context-dependent. Existing explainability assessments vary widely, spanning qualitative user studies and quantitative metrics like faithfulness and plausibility, yet no universal consensus exists on the most effective approach.

To address these challenges, we introduce \textit{EvalxNLP}, a benchmarking framework for evaluating post-hoc explainability methods in text classification tasks. \textit{EvalxNLP} supports multiple explanation techniques,
% including LIME \cite{ribeiro2016whyitrustyou}, SHAP \cite{lundberg2017unifiedapproachinterpretingmodel}, and Integrated Gradients \cite{sundararajan2017axiomaticattributiondeepnetworks},
and assesses them across key properties such as faithfulness, plausibility, and complexity. In addition, \textit{EvalxNLP} integrates LLM-based natural language explanations to facilitate the users' understanding of the generated explanations and evaluations. We also conduct a user-based study to evaluate the usability and user satisfaction with the framework. 
Our framework provides a systematic, user-friendly platform that aims to democratize access to explainability tools, enabling both researchers and practitioners to compare and refine XAI techniques for NLP applications. By offering a unified and reproducible evaluation methodology, \textit{EvalxNLP} advances the field of explainability, promoting more transparent and trustworthy AI systems. 

% these new lines are added because I had to remove some lines reg copyright from the cls file so we are doing this to maintain the same format similar to the one that will be published:
\newpage

\section{Related Work}
Most existing explainability evaluation frameworks are designed for general-purpose applications, meaning that they include explainability methods used in image or tabular applications. As a result, most of them lack dedicated support for text-based models. OpenXAI \cite{agarwal2022openxai}, BEExAI \cite{sithakoul2024beexai}, and Quantus \cite{hedstrom2023quantus} are prime examples of supporting multiple data modalities but not including text-specific evaluation metrics. Performance assessment is typically based on generic criteria, without tailored adaptations for NLP tasks.

Frameworks such as Inseq \cite{Sarti_2023_inseq} support sequence generation models but lack built-in evaluation metrics. XAI-Bench \cite{liu2021syntheticbenchmarksscientificresearch} evaluates explainability methods using synthetic data, which may not fully capture real-world text applications \cite{Faber2021GroundTruth}. While these frameworks provide partial solutions, they do not offer a comprehensive suite for text explainability. ferret \cite{attanasio-etal-2023-ferret} facilitates explainability evaluation for NLP models but supports only five feature-attribution methods and six evaluation metrics. Among existing XAI libraries, it is the only one with adequate text-specific explainability features, to be considered as an NLP-specialized explainability framework. But, it relies on some metrics that have been shown to be inaccurate especially for measuring faithfulness (as explained in section \ref{sec:faithfulness}); Captum \cite{miglani-etal-2023-captum} provides implementations for 22 attribution methods but is limited to two evaluation metrics (Infidelity and Sensitivity) and lacks built-in benchmarking. Similarly, AIX360 \cite{arya2022ai} supports text data but evaluates only two properties, faithfulness and monotonicity, without systematic benchmarking capabilities. M4 \cite{li2023M4} evaluates faithfulness for image and text modalities but does not assess plausibility or complexity.

While these frameworks address various aspects of explainability, among existing XAI libraries, only ferret offers a complete set of key features \cite{attanasio-etal-2023-ferret}: multiple explainability methods, Transformers-readiness (built with close integration into the Hugging Face (HF) \textit{transformers} library), evaluation APIs, explainable datasets (i.e., those with human-annotated rationales), and built-in visualization. In addition to these features, our tool also extends functionality by incorporating recent explainability methods, recent metrics for evaluating explanation properties and providing an LLM-based module that generates natural language explanations to enhance user understanding. It consolidates capabilities scattered across multiple frameworks, offering a robust suite for benchmarking, evaluation, and qualitative explanations. By seamlessly integrating these features into one comprehensive framework, like \textit{EvalxNLP},  we support practitioners in benchmarking Ph-FA explanation methods.
 
% Based on the comparison in \cite{attanasio-etal-2023-ferret}, among existing XAI libraries, only ferret offers a complete set of key features: multiple explainability methods, Transformers-readiness (supporting transformer-based models from Hugging Face(HF)), evaluation APIs, explainable datasets (i.e., those with human-annotated rationales), and built-in visualization. In addition to these features, our tool also extends functionality by incorporating recent explainability methods, recent metrics for evaluating explanation properties and providing an LLM-based module that generates natural language explanations to enhance user understanding.

% To the best of our knowledge, it is the first and most complete framework to seamlessly combine text-based evaluation metrics, feature attribution methods, and natural language explanations within a single, comprehensive package.

\section{The Framework}
% \textbf{ToDo: Describe the integration with HF}\\
% \textbf{Todo: describe the transformer readiness part}\\
% - Figure as an overview of the tool showing different components. 
% - In the metrics emphasize on using new metrics compared to old ones and show their criticism. 
% - show one example of code snippet. 
% Goals?
% Users? 
% etc. ACL checklist?
% focus on HF and extensions...
        
    The \textit{EvalxNLP} framework builds on the top of four main components described below. For the technical implementation details, we refer the reader to the documentation provided in the repository. We release the code, tutorials, and documentation in the following repository\footnote{\url{https://github.com/dmah10/EvalxNLP}}.

% Based on the comparison in \cite{attanasio-etal-2023-ferret}, among existing XAI libraries, only ferret offers a complete set of key features: multiple explainability methods, Transformers-readiness (supporting transformer-based models from Hugging Face(HF)), evaluation APIs, explainable datasets (i.e., those with human-annotated rationales), and built-in visualization. In addition to these features, our tool also extends functionality by incorporating recent explainability methods, recent metrics for evaluating explanation properties and providing an LLM-based module that generates natural language explanations to enhance user understanding.

\begin{table}

  \caption{Overview of the built-in explanation methods, metrics, and datasets in EvalxNLP}
  \label{tab:overview}
  \scriptsize
  \begin{tabular}{ll|ll|ll}
    \toprule
    Method Category & Explainers & Property & Metric & Dataset & Classification Task \\
    \midrule
    Gradient-based & Saliency & Faithfulness & Soft suff & MovieReviews & Sentiment \\
                   & GradientxInput &   & Softcomp & HateXplain & Hate Speech\\
                   & IntegratedGradients &   & FAD N-AUC & e-SNLI & NLI \\
                   & DeepLIFT &   & AUC-TP &  \\
    Perturbation-based & LIME & Plausibility & IOU-F1 & \\
                       & Guided BackProp &   & Token-F1 & \\
                       & SHAPIQ &    & AUPRC & \\
                       & SHAP & Complexity & Complexity & \\
                       &  &   & Sparseness &\\
                   
    \bottomrule
  \end{tabular}
\end{table}

% \begin{table}
%   \caption{Overview of the built-in explanation methods, metrics, and datasets in \textit{EvalxNLP}}
%   \label{tab:overview_\textit{EvalxNLP}}
%   \scriptsize
%   \begin{tabular}{
%     >{\columncolor{lightgray1}}l
%     >{\columncolor{lightgray1}}l |
%     >{\columncolor{lightgray2}}l
%     >{\columncolor{lightgray2}}l |
%     >{\columncolor{lightgray3}}l
%     >{\columncolor{lightgray3}}l
%   }
%     \toprule
%      Method Category & Explainers & Property & Metric & Dataset & Classification Task \\
%     \midrule
%     Gradient-based & Saliency & Faithfulness & Soft suff & MovieReviews & Sentiment \\
%                    & GradientxInput &   & Softcomp & HateXplain & Hate Speech\\
%                    & IntegratedGradients &   & FAD N-AUC & e-SNLI & NLI \\
%                    & DeepLIFT &   & AUC-TP &  &\\
%     Perturbation-based & LIME & Plausibility & IOU-F1 & &\\
%                        & Guided BackProp &   & Token-F1 & & \\
%                        & SHAPIQ &    & AUPRC & &\\
%                        & SHAP & Complexity & Complexity & & \\
%                        &  &   & Sparseness & &\\
%     \bottomrule
%   \end{tabular}
% \end{table}

\subsection{Explainers}
One goal of \textit{EvalxNLP} is to enable users to generate diverse explanations through multiple explainability methods. The explainer component integrates eight widely recognized explainability methods from the XAI literature, specifically focusing on post-hoc feature attribution (Ph-FA) methods. \textit{EvalxNLP} incorporates two categories of post-hoc methods: gradient-based and perturbation-based approaches. Gradient-based methods compute feature importance by leveraging gradients of the model’s output with respect to its input features. They efficiently utilize backpropagation, making them well-suited for deep learning models. \textit{EvalxNLP} integrates five key methods, including Saliency (also called Gradients)\cite{simonyan2014deepinsideconvolutionalnetworks}, which calculates raw gradients to highlight important inputs; Gradient×Input \cite{shrikumar2017justblackboxlearning}, which scales gradients by input values for enhanced clarity; Integrated Gradients \cite{sundararajan2017axiomaticattributiondeepnetworks}, which averages gradients along a path from a baseline to the input; DeepLift \cite{shrikumar2019learningimportantfeaturespropagating}, which attributes differences in activations to individual inputs for more stable attributions; and Guided BackProp \cite{springenberg2015strivingsimplicityconvolutionalnet}, which filters negative gradients to highlight only positive contributions. \textit{EvalxNLP} implements these methods using Captum and ferret, providing a diverse set of efficient and interpretable attribution techniques.. Perturbation-based methods integrated into the toolbox include the widely used LIME \cite{ribeiro2016whyitrustyou} and SHAP \cite{lundberg2017unifiedapproachinterpretingmodel} methods, as well as the recently introduced SHAP with interactions method (SHAP-I) \cite{muschalik2024shapiqshapleyinteractionsmachine}, which augments traditional Shapley values by incorporating feature interactions \cite{shapiq}, a notable contribution over existing frameworks. The rationale behind providing a comprehensive range of Ph-FA methods is twofold: (1) to offer users access to a diverse set of explanations from established as well as novel methods, enabling a holistic assessment and selection of explanations tailored to specific use cases; and (2) to facilitate benchmarking, comparative analyses, and evaluations of these methods based on selected evaluation criteria.

 We implement the explanation methods by building on top of their original implementations (for LIME and SHAP) and existing open-source libraries for the remaining methods. Specifically, we use the original implementation for LIME. For SHAP, we utilize Partition SHAP, a variant that optimizes Shapley value computation by exploiting feature independence. For SHAP-I, we extend the implementation provided by the shapiq package \cite{muschalik2024shapiq}, while gradient-based methods are implemented using the Captum library \cite{miglani-etal-2023-captum}. Our approach of building upon established open-source libraries aims to facilitate and support the expansion and development of open-source XAI libraries. 

\subsection{LLM explanations}
% \textbf{todo:} motivation: refer to some papers that talk about the relevance/usefulness of NLEs by LLMs.

\begin{figure}[ht]
    \centering
    \begin{subfigure}{0.3\textwidth}
        \centering
           \includegraphics[scale=0.19]{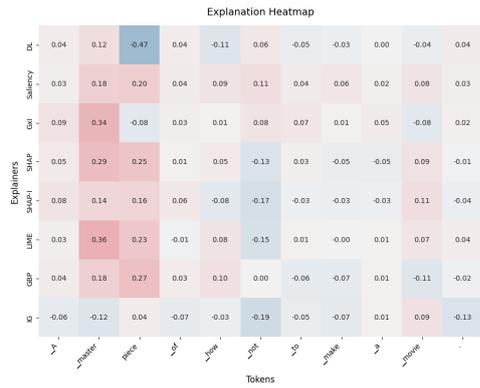}
        \caption{\label{fig:heatmap-misclassified}Explanation heatmap}
    \end{subfigure}
    \hfill
    \begin{subfigure}{0.54\textwidth}
        \centering
        \includegraphics[height=0.2\textheight, width=\textwidth]
        {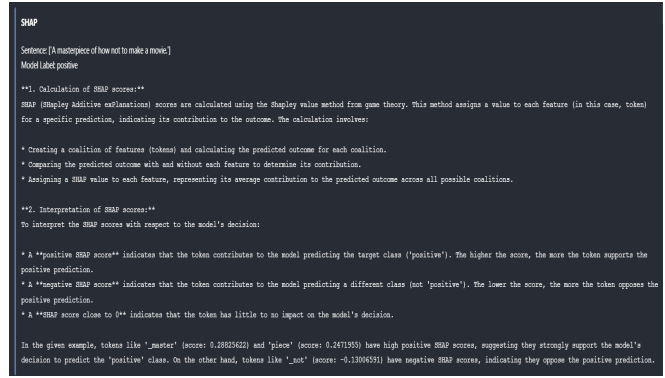}
        % {figures/shap_llm_explanation.pdf}
         % \includegraphics[scale=0.5]{figures/shap_llm_explanation.pdf}
        \caption{\label{fig:llm-shap-misclassified}LLM-generated explanation}
        
    \end{subfigure}
    \caption{\label{fig:heatmap-and-llm-explanation}(\textbf{Zoom in for a better view)}: (a) heatmap of the importance scores generated by the explainers for the single misclassified instance \textit{A masterpiece of how not to make a movie} and (b) LLM-generated textual explanation for the scores by SHAP}

\end{figure}

% Due to the capabilities of LLMs in generating ....the usage of LLMs for explaining...
Feature attribution explanations, especially those involving many features, can be difficult for lay users to interpret \cite{feldhus-saliency-verbalization-2023}. To address this, our tool integrates an LLM-based module that automatically generates natural language explanations to help users interpret both (1) the importance scores from various explanation methods and (2) the evaluation metric scores. These textual explanations enhance the comprehensibility of Ph-FA outputs, particularly for non-experts, and support more informed decision-making by providing textual explanations for the evaluation metric scores Additionally, combining visual heatmaps with textual explanations offers a more accessible and holistic view of the model’s decision-making process. To prevent the unfaithful textual explanations for model decisions by LLMs
\cite{feldhus-saliency-verbalization-2023},
% \cite{agarwal2024faithfulnessvsplausibilityunreliability},
we don't ask the LLM to provide its own explanations of the model decisions but only use LLM solely to verbalize the outputs of explanation methods such as importance scores into natural language to make them more comprehensible. Figure \ref{fig:heatmap-and-llm-explanation} presents an example of an explanation heatmap (Figure \ref{fig:heatmap-misclassified}) and LLM-generated explanation for the SHAP scores (Figure \ref{fig:llm-shap-misclassified}) for a misclassified instance in the MovieReviews dataset \cite{Zaidan2008ModelingAA} that is misclassified by XLM-RoBERTa-base \cite{barbieri-xlm-roberta-2021}. As shown in Figure \ref{fig:llm-shap-misclassified}, the LLM provides comprehensible textual explanation to ease understanding the scores by SHAP. 

The LLM is integrated into the framework via an API, enabling seamless generation of textual explanations on demand. For LLM API support, our demo uses the Together AI\footnote{\url{https://www.together.ai/}} API, 
with Llama-3.3-70B-Instruct-Turbo as the default model for generating explanations. Users can switch models or providers and modify the LLM instructions as needed.
% with Llama-3.3-70B-Instruct-Turbo as the default model for generating textual explanations. Users can easily switch to a different model for the same provider. The module also allows users to change the API provider by modifying the provider to suit their needs. The instructions to the LLM can be easily modified by the user when needed.

\subsection{Metrics}

To evaluate explainability methods within our NLP framework, we use a comprehensive set of metrics covering three key properties: faithfulness, plausibility, and complexity. These ensure that explanations align with model reasoning while remaining interpretable and concise for users. By integrating diverse metrics, the framework supports a rigorous, holistic assessment that balances model fidelity, human interpretability, and explanation brevity. For mathematical details, we refer readers to the original papers introducing these metrics. Below, (↓)/(↑) indicates that lower/higher values are better for a given metric.

\subsubsection{Faithfulness} \label{sec:faithfulness}
Faithfulness measures how well the generated explanations reflect the true behavior of the model.
Compared to other frameworks that use sufficiency (suff) and comprehensiveness (comp) based on complete token removal, an approach shown to produce inaccurate faithfulness measurements \citep{aopc_unfaithfulness_2022}, we employ soft suff and soft comp, which have proven more accurate in measuring faithfulness \citep{aopc-limitations-2023}. 
\textbf{Soft suff} \cite{zhao-aletras-2023-incorporating} ↓  : Evaluates how well the most important tokens can retain the model’s prediction when other tokens are softly perturbed. It assumes that retaining more elements of important tokens should preserve the model's output, while dropping less important tokens should have minimal impact.
% The suff score is computed based on the difference in prediction probability.
\textbf{Soft comp} \cite{zhao-aletras-2023-incorporating} ↑ : Measures how much the model's prediction changes when important tokens are softly perturbed using Bernoulli mask. It assumes that heavily perturbing important tokens should significantly affect the model's output, indicating their importance to the prediction.
% The comp score is computed as the difference between the original and perturbed prediction probabilities.
\textbf{Feature Attribution Dropping (FAD) Curve and Normalized Area Under Curve (N-AUC)} \cite{ngai-rudzicz-2022-doctor} ↓ : Measures the impact of dropping the most salient tokens on model performance, with the steepness of the FAD curve indicating the method’s faithfulness. The N-AUC quantifies this steepness, where a lower score reflects better alignment of the attribution method with the model’s true feature importance.
\textbf{Area Under the Threshold-Performance Curve (AUC-TP)} \cite{atanasova2020diagnosticstudyexplainabilitytechniques} ↓ : AUC-TP evaluates the faithfulness of saliency explanations by progressively masking the most important tokens (based on their saliency scores) and measuring the drop in the model’s performance. This AUC-TP value provides a single metric summarizing how significantly the model relies on the highlighted tokens, with lower values indicating better faithful explanations.

\subsubsection{Plausibility}
Plausibility measures how well the generated explanations align with human intuition. The following metrics are incorporated: \textbf{IOU-F1 Score} \cite{deyoung-etal-2020-eraser} ↑ : Computes the Intersection over Union (IoU) between predicted and ground-truth rationales, considering a (partial) match if the overlap is above 50\% where these matches are used to calculate F1 scores. \textbf{Token-Level F1 Score} \cite{deyoung-etal-2020-eraser} ↑ : Measures alignment by calculating the F1-score between predicted and human-annotated rationales at the token level. \textbf{Area Under Precision-Recall Curve (AUPRC)} \cite{deyoung-etal-2020-eraser} ↑ : Evaluates plausibility by comparing the saliency scores of tokens with ground-truth rationale masks, computing the area under the precision-recall curve.
% \begin{itemize}
%     \item \textbf{IOU-F1 Score} \cite{deyoung-etal-2020-eraser} ↑ : Computes the Intersection over Union (IoU) between predicted and ground-truth rationales, considering a match if the overlap is above 50\%.
%     \item \textbf{Token-Level F1 Score} \cite{deyoung-etal-2020-eraser} ↑ : Measures alignment by calculating the F1-score between predicted and human-annotated rationales at the token level.
%     \item \textbf{AUPRC} \cite{deyoung-etal-2020-eraser} ↑ : Area Under Precision-Recall Curve (AUPRC) evaluates plausibility by comparing the saliency scores of tokens with ground-truth rationale masks, computing the area under the precision-recall curve.
% \end{itemize}

\subsubsection{Complexity}
Complexity evaluates how concise and interpretable the explanations are. Sparse explanations that highlight only a few important features are preferred. We use the following metrics:
% \begin{itemize}
%     \item \textbf{Complexity} \cite{bhatt2020evaluatingaggregatingfeaturebasedmodel} ↓ : Measures how evenly importance scores are distributed using Shannon entropy. Higher values indicate more complex explanations, while lower values suggest more concise attributions.
%     \item \textbf{Sparseness} \cite{chalasani2020conciseexplanationsneuralnetworks} ↑ : Computes the sparsity of attributions using the Gini index, where higher scores indicate more concentrated importance on a few features.
% \end{itemize}
\textbf{Complexity} \cite{bhatt2020evaluatingaggregatingfeaturebasedmodel} ↓ : Measures how evenly importance scores are distributed using Shannon entropy. Higher values indicate more complex explanations, while lower ones suggest concise attributions. 
\textbf{Sparseness} \cite{chalasani2020conciseexplanationsneuralnetworks} ↑ : Computes the sparsity of attributions using the Gini index, where higher scores indicate more concentrated importance on a few features.

\subsection{Datasets}
\textit{EvalxNLP}
% is a comprehensive framework designed for text classification tasks, offering robust evaluation and explanation capabilities for NLP models. It supports key applications such as Sentiment Analysis, which classifies text sentiment as positive, negative, or neutral, commonly used in product reviews, social media, and customer feedback analysis. It also includes Hate Speech Detection, identifying offensive or harmful content to help moderate online platforms. Additionally, Natural Language Inference (NLI) assesses the logical relationship between a premise and a hypothesis, determining entailment, contradiction, or neutrality, crucial for applications like question answering and text summarization. For each of these tasks, our framework includes rationale-annotated datasets (also called explainable datasets), incorporating human-annotated rationales that highlight the most critical words or sentences for a given class label. These rationales enable an evaluation of the alignment between model-generated explanations and human understanding. 
% The datasets are naturally a by product of the NLP tasks that are supported in the framework, therefore we included
is a framework for text classification that combines robust evaluation and explanation capabilities. It supports tasks like Sentiment Analysis, Hate Speech Detection, and Natural Language Inference (NLI), using rationale-annotated datasets that highlight key text segments. These human-provided rationales enable assessment of how well model explanations align with human reasoning. Currently, \textit{EvalxNLP} includes a representative dataset for each of the aforementioned tasks and supports these three datasets by default, while also allowing users to extend it with additional classification datasets:
% \begin{itemize}
%     \item \textbf{MovieReviews}: Designed for Sentiment Analysis, this dataset consists of 1,000 positive and 1,000 negative movie reviews, each annotated with phrase-level human rationales that justify the sentiment label.
%     \item \textbf{HateXplain} \cite{mathew2022hatexplainbenchmarkdatasetexplainable}: Used for Hate Speech Detection, this dataset comprises 20,000 posts from Gab and Twitter, annotated with one of three labels: hate speech, offensive, or normal.
%     \item \textbf{e-SNLI} \cite{camburu2018esnlinaturallanguageinference}: A dataset for Natural Language Inference containing 549,367 examples, split into training, validation, and test sets. Each example includes a premise and a hypothesis labeled as entailment, contradiction, or neutral.
% \end{itemize} 
\textbf{MovieReviews}: Designed for Sentiment Analysis, this dataset consists of 1,000 positive and 1,000 negative movie reviews, each annotated with phrase-level human rationales that justify the sentiment label. \textbf{HateXplain} \cite{mathew2022hatexplainbenchmarkdatasetexplainable}: Used for Hate Speech Detection, this dataset comprises 20,000 posts from Gab and Twitter, annotated with one of three labels: hate speech, offensive, or normal. \textbf{e-SNLI} \cite{camburu2018esnlinaturallanguageinference}: A dataset for Natural Language Inference containing 549,367 examples, split into training, validation, and test sets. Each example includes a premise and a hypothesis labeled as entailment, contradiction, or neutral.

% incorporating timely classification tasks such as Fake News Detection and Intellectual Property Classification. Fake News Detection leverages linguistic patterns and factual verification to identify misleading or false information, playing a critical role in combating misinformation across digital platforms. Meanwhile, Intellectual Property Classification aids in categorizing and protecting proprietary content by distinguishing between copyrighted, trademarked, or publicly available materials. Expanding the framework with these capabilities enhances its versatility, making it a more comprehensive solution for content moderation, legal compliance, and information integrity in an ever expanding digital world.

\section{Case Study}
To show the usability of \textit{EvalxNLP} on real-world datasets, we present a case study showcasing how the tool can be used for benchmarking explainers for a sentiment analysis task on the Movie Reviews that include rationales \cite{deyoung-etal-2020-eraser} using an XLM-RoBERTa-base that is fine-tuned for sentiment analysis. \textit{EvalxNLP} enables users to generate explanations and benchmark explainers either for single instances (local explanations) and across multiple instances (a subset or an entire dataset). This functionality depends on the user's intention, such as identifying the best explanation method with respect to specific metrics and properties, either for individual sentences or aggregated across datasets. In this case study, we demonstrate how \textit{EvalxNLP} benchmarks explainers using the full Movie Reviews dataset by aggregating evaluation metrics across all instances. 

% \begin{table*}[ht]
% \caption{Evaluation metrics results for the explainers on the Movie Reviews dataset with best scores in bold.}
% \label{tab:case_study}
% \scriptsize
% \centering
% % \begin{adjustbox}{width=\textwidth}
% \begin{tabular}{lccccccccc}
% \toprule
% \textbf{Methods} & \textbf{IG} & \textbf{GBP} & \textbf{DL} & \textbf{GxI} & \textbf{Saliency} & \textbf{LIME} & \textbf{SHAP} & \textbf{SHAP-I} \\
% \midrule
% Softcomp $\uparrow$ & 0.1160 & 0.0963 & 0.0950 & 0.1194 & \textbf{0.2506} & 0.0888 & 0.1184 & \textbf{0.1846} \\
% Soft suff $\downarrow$ & 0.0091 & 0.0132 & 0.0180 & 0.0142 & \textbf{0.0015} & \textbf{0.0022} & 0.0473 & 0.0911 \\
% FAD $\downarrow$ & \textbf{0.8908} & 0.9055 & \textbf{0.8824} & 0.9391 & 0.9559 & 0.9202 & 0.9475 & 0.9370 \\
% AUTPC $\downarrow$ & 0.5075 & 0.5030 & 0.5101 & 0.4970 & 0.5010 & \textbf{0.4879} & 0.5151 & \textbf{0.4915} \\
% Complexity $\downarrow$ & \textbf{0.0113} & 0.0116 & 0.0115 & 0.0115 & 0.0121 & 0.0118 & 0.0117 & 0.0120 \\
% Sparseness $\uparrow$ &\textbf{0.5149} & 0.4921 & 0.4922 & \textbf{0.5078} & 0.2925 & 0.4231 & 0.4255 & 0.2851 \\
% IOU F1 score $\uparrow$ & 0.1855 & \textbf{0.1880} & 0.1866 & 0.1845 & 0.1857 & 0.1853 & \textbf{0.2057} & 0.1877 \\
% Token F1 score $\uparrow$ & 0.2940 & \textbf{0.2974} & 0.2953 & 0.2927 & 0.2945 & 0.2938 & \textbf{0.3158} & 0.2960 \\
% AUPRC $\uparrow$ & 0.2447 & 0.2497 & 0.2473 & 0.2501 & \textbf{0.2559} & 0.2489 & \textbf{0.2904} & 0.2480 \\
% \bottomrule
% \end{tabular}
% % \end{adjustbox}
% \end{table*}

\begin{figure}[t]
    \centering
    \includegraphics[width=1\textwidth]
    {figures/evaluation_case_study_framework.pdf}
    \caption{Evaluation metrics results from EvalxNLP for the explainers on the \textit{MovieReviews} dataset with best scores in \textbf{bold}. Darker colors refer to better values for each metric, and bold refers to the best value per metric. 
    % Metrics are color-grouped based on the property they measure from left to right.
    }
    \label{fig:evaluation_scores_case_study_EvalxNLP}
\end{figure}

Figure \ref{fig:evaluation_scores_case_study_EvalxNLP} presents the results indicating that DL achieves the highest overall faithfulness scores, particularly on soft metrics, suggesting it produces the most faithful explanations for this dataset. IG performs best on complexity metrics, indicating its explanations are simpler and easier to understand, while SHAP outperforms other methods on plausibility metrics, showing that its explanations align closely with human intuition. As expected, no single explanation method excels in all properties, confirming that practitioners must select methods based primarily on the evaluation property most relevant to their specific use case.

\section{Human Evaluation}
% Mahdi: maybe only add the evaluation on the framework and not the explainers (as we don't have enough space. ). Something like: we plan to add these results in an extended version of this paper e.g. for the camera-ready or separate extension.

% draft:
% We also performed a user-based study (human evaluation) to assess the usability of the tool...
% In the first part of the study, we measured the simulatability of the explanations,...
% In the second part, we asked user to use the tool and then collected their feedback on the following aspect using a likert-scale...

% other metrics: pre-selected examples from the datasets, ones that were 

% Usability of the tool: 

\begin{figure}[ht]
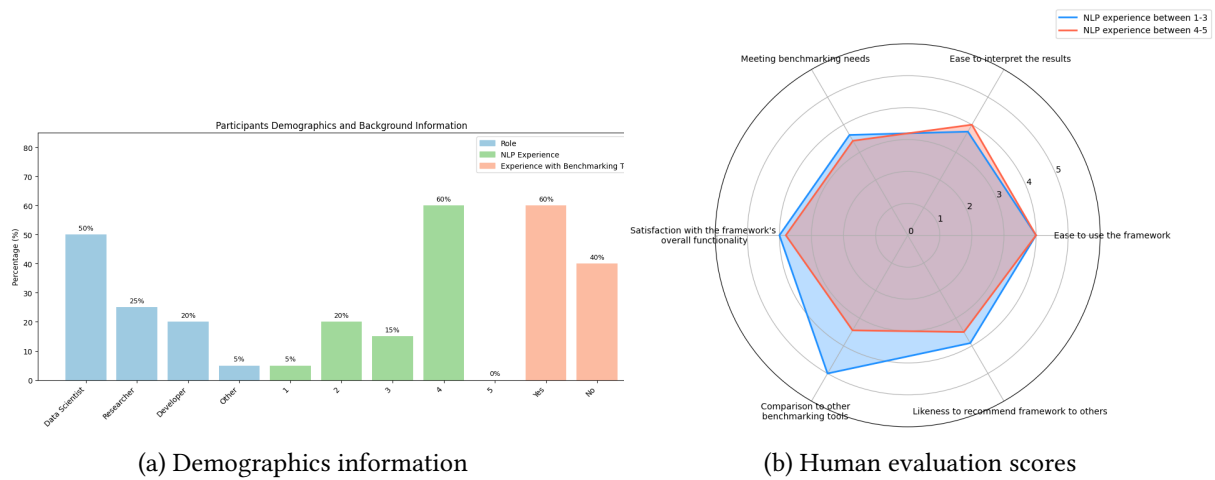

    \centering
    \begin{subfigure}{0.49\textwidth}
        \centering
           \includegraphics[scale=0.28]{figures/demographics.pdf} 
        \caption{\label{fig:demographics}Demographics information}
    \end{subfigure}
    \hfill
    \begin{subfigure}{0.49\textwidth}
        \centering
        \includegraphics[width=\textwidth]{figures/radar_chart_final.pdf}
        \caption{\label{fig:human_evaluation}Human evaluation scores}
        
    \end{subfigure}
    % \caption{Overall caption for the figure block (optional).}
    
    \caption{\label{fig:demographics-and-evaluation-scores} \textbf{Zoom in for a better view)} (a) Bar chart for the demographics and background information of the participants in the human evaluation. (b) Radar chart presenting the results of the human evaluation. Responses on framework comparison were collected only from participants who confirmed prior experience using other tools (2 participants for lower NLP experience and 9 participants for higher NLP experience).}
\end{figure}

We also conducted a user-based study involving 20 participants to assess the usability of the tool. The participants were provided with instructions on how to run the tool and try its different functionalities. The user study first collects demographic data, including participants' profession, NLP experience, and prior exposure to benchmarking tools. Then, they were asked to evaluate the system based on some criteria using a 5-point Likert scale to measure usability and satisfaction, where 1 and 5 mean the worst and best values, respectively, for each criterion (For brevity, we don't present the questions here and refer to Figure \ref{fig:demographics-and-evaluation-scores}). Figure \ref{fig:demographics-and-evaluation-scores} presents the results of the human evaluation where these results were collected prior to the integration of the LLM component, which was later added to enhance the understanding of explanations generated by the various explainers. 
Another round of human evaluation is planned to assess the effectiveness of adding the LLM-component. 

% \begin{figure}[ht]
%     \centering
%     % \includegraphics[width=0.9\textwidth]{figures/demographics.pdf}
%     \includegraphics[scale=0.3]{figures/demographics.pdf} 
%     \caption{Participants demographics and background information}
%     \label{fig:evaluation}
% \end{figure}

% \begin{figure}[htbp]

Based on results in Figure \ref{fig:human_evaluation}, 
 The overall results are promising, with scores consistently above 3 out of 5 across all criteria for both participant groups. Particularly, the framework is easy to use, and all users, especially those with greater NLP experience, find the results easy to interpret. However, for the remaining criteria, participants with less NLP experience provided higher ratings compared to their more experienced counterparts. This indicates there remains room for improvement, particularly in enhancing the framework's ability to meet the benchmarking needs of more experienced users.
 % \footnote{\textit{Due to space constraints, we present only summary results and will include a detailed analysis in the camera-ready version.}}

 % \textit{For brevity and space limitations,  we provide these results only, but we plan to present the detailed results and in-depth analysis of the human evaluation we conducted in the camera-ready version.}

\section{Conclusion}
Given the increasing number of available explainability methods and the diverse requirements stakeholders may have, there is a rising need for continued contributions to existing and new frameworks that support stakeholders in obtaining and selecting appropriate explanations tailored to their specific use cases. To address this gap, we introduce \textit{EvalxNLP}, a novel Python framework designed to benchmark state-of-the-art Ph-FA explainability methods for transformer-based NLP models, particularly targeting classification tasks. The framework enables users to generate and evaluate explanations at the single-instance level and across entire real-world datasets across different metrics for three main explainability properties: faithfulness, complexity, and plausibility. \textit{EvalxNLP} is targeted for use by various stakeholders, including laypeople, developers, and researchers, depending on their goals {  and also where certain properties could be more critical for specific users. 
% For example, developers can use \textit{EvalxNLP} to debug models and to compare the Ph-FA methods, focusing on faithfulness metrics due to its relevance for this user group.
For instance, developers can employ EvalxNLP to debug models and compare Ph-FA methods, prioritizing faithfulness metrics relevant to their needs.} Our framework is developed to be easily extensible by the research community. \textbf{Limitations} of the framework include focusing on classification tasks and utilizing feature-attribution explainability methods. \textbf{Future directions} include expanding the supported methods and metrics, such as integrating recent non-feature attribution techniques like \cite{slalom-2025} and robustness metrics such as sensitivity \cite{sensitivity-infidelity-2019}. 
% Future work also includes
We also plan to incorporate users feedback to
refine explanation quality, and to extend the framework to generate premise–conclusion rules from Ph-FA methods to enhance explainability\cite{rizzo-2024-rule-xai}.

\begin{acknowledgments}
We would like to thank the anonymous reviewers
for their helpful suggestions. This research has
been supported by the German Federal Ministry of
Education and Research (BMBF) grant 01IS23069
Software Campus 3.0 (TU München).  
\end{acknowledgments}

% %% The declaration on generative AI comes in effect
% %% in Janary 2025. See also
% %% https://ceur-ws.org/GenAI/Policy.html
% % \section*{Declaration on Generative AI}
% %   {\em Either:}\newline
% %   The author(s) have not employed any Generative AI tools.
% %   \newline
  
% %  \noindent{\em Or (by using the activity taxonomy in ceur-ws.org/genai-tax.html):\newline}
% %  During the preparation of this work, the author(s) used X-GPT-4 and Gramby in order to: Grammar and spelling check. Further, the author(s) used X-AI-IMG for figures 3 and 4 in order to: Generate images. After using these tool(s)/service(s), the author(s) reviewed and edited the content as needed and take(s) full responsibility for the publication’s content. 

% %%
% %% Define the bibliography file to be used
% % DO NOT CHANGE THE STYLE BELOW
% \bibliography{sample-ceur}
\bibliography{concise_references}

% %%
% %% If your work has an appendix, this is the place to put it.
% % \appendix

% % \section{Online Resources}

% % The sources for the ceur-art style are available via
% % \begin{itemize}
% % \item \href{https://github.com/xxx}{GitHub}
% % \end{itemize}

\end{document}